\pdfoutput=1

\documentclass{article}

\usepackage[preprint]{neurips_2019}

\usepackage[utf8]{inputenc} 
\usepackage[T1]{fontenc}    
\usepackage{hyperref}      
\usepackage{url}            
\usepackage{booktabs}       
\usepackage{amsfonts}       
\usepackage{nicefrac}       
\usepackage{microtype}  

\usepackage{algorithmic}

\usepackage{amsmath,amssymb,amsfonts}
\usepackage{graphicx}
\usepackage{graphics}
\usepackage{multirow} 
\usepackage{booktabs} 
\usepackage{amssymb} 
\usepackage[table]{xcolor}
\usepackage{colortbl}
\usepackage{hhline}
\usepackage{appendix}
\usepackage{caption}
\usepackage{subcaption}

\usepackage{amssymb,amsmath}
\usepackage[ruled,vlined]{algorithm2e}
\usepackage{enumitem}
\usepackage{fancyhdr}
\usepackage{xcolor}
\setlist[enumerate]{label*=\arabic*.}
\usepackage{mathtools}
\DeclarePairedDelimiterX{\infdivx}[2]{(}{)}{
  #1\;\delimsize\|\;#2
}

\title{SoQal: Selective Oracle Questioning in Active Learning}

\author{
 Dani Kiyasseh\\
 Department of Engineering Science\\
 University of Oxford\\
 Oxford, UK \\
 \texttt{dani.kiyasseh@eng.ox.ac.uk} \\
   \And
   Tingting Zhu\thanks{equal contribution} \\
   Department of Engineering Science\\
   University of Oxford\\
   Oxford, UK \\
   \texttt{tingting.zhu@eng.ox.ac.uk} \\
   \And
   David A. Clifton\footnotemark[1] \\
   Department of Engineering Science\\
   University of Oxford\\
   Oxford, UK \\
   \texttt{david.clifton@eng.ox.ac.uk} \\
}

\begin{document}

\maketitle

\begin{abstract}
  
Large sets of unlabelled data within the healthcare domain remain underutilized. Active learning offers a way to exploit these datasets by iteratively requesting an oracle (e.g. medical professional) to label instances. This process, which can be costly and time-consuming is overly-dependent upon an oracle. To alleviate this burden, we propose SoQal, a questioning strategy that dynamically determines when a label should be requested from an oracle. We perform experiments on five publically-available datasets and illustrate SoQal's superiority relative to baseline approaches, including its ability to reduce oracle label requests by up to 35\%. SoQal also performs competitively in the presence of label noise: a scenario that simulates clinicians' uncertain diagnoses when faced with difficult classification tasks. 

\end{abstract}

\section{Introduction}

The success of modern-day deep learning algorithms in the medical domain has been contigent upon the availability of large, labelled datasets \citep{Poplin2018,Tomasev2019,Attia2019}. The curation of such datasets, however, is a challenge due to the time-consuming nature of and high costs associated with labelling. This is particularly the case in the medical domain where the input of expert medical professionals is required. One way of overcoming this challenge and exploiting large, \textit{unlabelled} datasets is via active learning (AL) \citep{Settles2009}. In this setting, a learner is tasked with iteratively acquiring a subset of unlabelled instances and asking an oracle to label it, before adding it to the set of labelled instances. By presenting the most informative instances to the oracle, AL aims to improve the performance of algorithms while minimizing the burden of labelling on the oracle. 

Although shown to be data-efficient, current AL approaches are overly \textit{reliant} on the presence of an oracle. Namely, an oracle is always assumed to be present. Such over-reliance is detrimental for two reasons. Firstly, it negatively affects the applicability of AL algorithms to scenarios where an oracle is either unavailable or is ill-trained for the task at hand. This is prevalent, for instance, in low-resource healthcare settings where there is a shortage of qualified medical professionals. Secondly, over-reliance can still inundate experts with a significant number of label requests, the very goal AL is supposed to minimize. This is particularly consequential for expert medical professionals who have limited bandwidth and who are increasingly suffering from 'burnout' \citep{West2016,Shanafelt2017}.

Decreasing the dependence of AL algorithms on the oracle and further alleviating their associated labelling burden can significantly improve the manner in which a physician is involved while also overcoming limitations inherent in the oracle. Although \cite{Kiyasseh2020} suggest performing oracle-free active learning, they only consider the extreme scenarios where an oracle is either unavailable or always available. We hypothesize that an algorithm capable of finding a middle-ground in terms of an oracle strategy could lead to less dependence on an oracle while not compromising, or potentially improving, performance. To design such an algorithm, we take inspiration from work in selective classification \citep{Chow1970,El-Yaniv2010} where algorithms learn to abstain from making a prediction. 

\textbf{Our Contributions.} In this paper, we challenge the traditional assumptions of active learning, namely the availability of noise-free oracles and propose a dynamic strategy to deal with this. 
\begin{enumerate}
\item \textbf{Selective Oracle Questioning (SoQal)}: a dynamic strategy that learns when to request a label from an oracle during active learning. 
\item A novel objective function that helps a network predict the zero-one classification loss incurred on the main task. We use this prediction to control the dependence of the network on an oracle. 
\end{enumerate} 

\section{Related Work}

\textbf{Active learning and healthcare} have been relatively under-explored. A recent review of active learning methodologies can be found in \citet{Settles2009}. In the healthcare domain, \citet{Gong2019} propose to acquire instances from an electronic health record (EHR) database using a Bayesian deep latent Gaussian model to improve mortality prediction. \citet{Smailagic2018, Smailagic2019} introduce MedAL, a method that actively acquires unannotated medical images by measuring their distance in a latent space to images in the training set. Such similarity metrics, however, are sensitive to the original amount of labelled training data. The work of \citet{Wang2019} is similar to ours in that they focus on the electrocardiogram. \citet{Gal} adopt BALD \citep{Houlsby2011} in the context of Monte Carlo Dropout to acquire datapoints that maximize the Jensen-Shannon divergence (JSD) across MC samples. There have been several attempts at learning from multiple or imperfect labelers \citep{Dekel2012,Zhang2015,Sinha2019}. \cite{Urner2012} propose choosing the oracle that should label a particular instance. Unlike our approach, they do not explore independence from an oracle. \cite{Yan2016} do consider abstention in an AL setting, yet it is performed by the labeler. Instead, our approach places the decision of abstention under the control of the learner. To the best of our knowledge, previous work, in contrast to ours, has assumed the existence of an oracle and has not explored a dynamic oracle selection strategy. 

\textbf{Selective classification and healthcare} fit well with one another given the high-stakes scenarios present in the latter. Early work in selective classification by \cite{Chow1970} introduces the risk-coverage trade-off whereby the empirical risk of a model is inversely related to its rate of abstentions. \citet{El-Yaniv2010} define perfect learning as an empirical risk of zero that corresponds to non-zero coverage of instances and propose the Consistent Selective Strategy (CSS) to achieve this. \citet{Wiener2011} use a support vector machine (SVM) to rank and reject instances based on the degree of disagreement between hypotheses. In some frameworks, these happen to be the same instances that active learning treats as being most informative. More recently, \citet{Cortes2016} outline an objective function that penalizes inappropriate abstentions alongside the rate at which they are performed. Instead of using uncertainty-based heuristics such as the entropy of the posterior predictive distribution or the Softmax-Response \citep{Geifman2017}. \citet{Ziyin2019} exploit portfolio theory and propose the gambler's loss in order to learn a selection function that determines whether instances are rejected. However, this approach requires a significant amount of hyperparameter tuning. Most similar to our work is SelectiveNet \citep{Geifman2019} where a multi-head neural architecture is used in conjunction with an empirical selective risk (ESR) objective function and a percentile threshold. In contrast, our work proposes a different objective function, a thresholding mechanism, and specifically considers oracle selection. Moreover, ESR assumes that the labels associated with instances are known. In contrast, we extend the idea of selective classification to the setting where labels are \textit{unknown}. 

\section{Methods}

\subsection{Active Learning}

In this work, we consider a learner $f_{\omega}: X \rightarrow Y$, a neural network parameterized by $\omega$ that maps inputs $X \in \mathbb{R}^m$ to outputs $Y \in [1\ldots\textit{C}]$, where \textit{C} is the number of classes. After training on a pool of labelled data $\textit{L} = (X_{L},Y_{L})$ for $\tau$ epochs, the learner is tasked with querying the unlabelled pool of data $U = (X_{U},Y_{U})$ and acquiring the top $b\%$ of instances, $x_{b} \sim X_{U}$, that it deems to be most informative. 

The degree of informativeness of an instance is determined by an acquisition function, $\alpha$, such as that found in Bayesian Active Learning by Disagreement (BALD) \citep{Houlsby2011} or ALPS \citep{Kiyasseh2020}. Such approaches when used in conjunction with Monte Carlo Dropout (MCD) \citep{Gal2016} identify instances that lie in the region of classification uncertainty. This is a region in which hypotheses disagree the most about instances. One forward pass of MCD outputs a softmax posterior distribution $p(y|x,\omega_{t})$ where $\omega_{t} \sim q_{\theta}(\omega)$ represents parameters sampled from the MC distribution. To obtain an accurate approximation of the hypothesis space, this is repeated \textit{T} times resulting in $G \in \mathbb{R}^{T\text{x}C}$ for each instance. 
\begin{equation}
\begin{split}
\mathrm{BALD_{MCD}} = \mathrm{JSD}(p_{1},p_{2},\ldots,p_{T}) 
&= \mathrm{H}(p(y|x)) - \mathbb{E}_{p(w|D_{train})} \left[\mathrm{H}(p(y|x,w)) \right] \\
&\approx \mathrm{H} \left (\frac{1}{T} \sum_{t=1}^{T} p(y|x,\omega_{t}) \right) - \frac{1}{T} \sum_{t=1}^{T} \left[\mathrm{H}(p(y|x,\omega_{t})) \right]
\end{split}
\label{eq:bald}
\end{equation}
where JSD is the Jensen-Shannon Divergence and H represents the entropy function. Once instances are acquired, they are provided to an oracle, who is assumed to be available and noise-free, for labelling before being added to the pool of labelled instances. This process is repeated until the performance of an algorithm is considered to be sufficient.

\subsection{Selective Oracle Questioning}
\label{section:soqal}

Traditionally, in AL, requesting a label from an oracle automatically follows the act of selecting an unlabelled instance. We challenge this convention and treat these two processes as independent of one another. This section describes how to choose whether or not to request a label \textit{after} an unlabelled instance has been chosen.

\begin{figure}[!h]
\centering
\begin{subfigure}[h]{\textwidth}
        \centering
        \includegraphics[width=\textwidth]{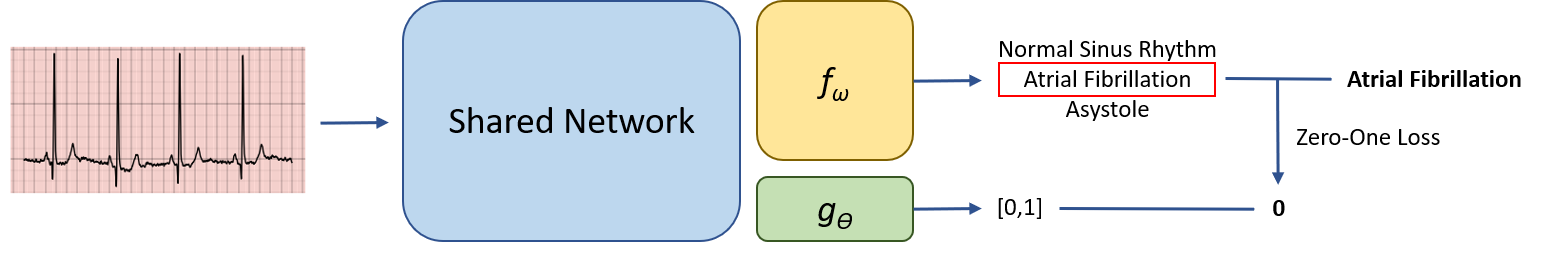}
\end{subfigure}
        \caption{Selective Oracle Questioning Framework.}
        \label{fig:network_architecture}
\end{figure}

\textbf{Architecture.} We assume the existence of a prediction network, $f_{\omega}$, which for each instance, $x$, generates posterior class probabilities, $p(y|x,\omega)$, and an oracle selection network, $g_{\theta}: X \rightarrow o \in [0,1]$ parameterized by $\theta$ that maps that same instance to a scalar, as shown in Fig.~\ref{fig:network_architecture}. 

\textbf{Objective Function.} We interpret the scalar, $o$, as approximating the probability that an oracle is requested for a label. Ideally, a network should only be reliant on an oracle when it cannot classify an instance correctly itself. Therefore, high values of $o$ should be associated with incorrect network predictions. Conversely, low values of $o$ should be associated with correct network predictions. We encourage this behaviour by assigning the zero-one loss, \textit{e}, of $f_{\omega}$ as the ground truth label for $g_{\theta}$. 

We note that, for each instance, this ground truth label will inevitably shift during training as the network becomes more adept at classifying it. Early in training, the ratio of misclassified to correctly classified instances will be high. Late in training, the opposite is true. If such ratios are left unaccounted for, with \textit{e} being used as the ground truth label, the majority of the outputs of $g_{\theta}$ will be high early during training and low near the end. Therefore, distinguishing between individual instances based solely on the output of $g_{\theta}$ would be difficult and thus deem it an unreliable signal for oracle selection. This scenario is equivalent to that of class imbalance. We describe how to mitigate this effect below. 

Our objective function for a mini-batch of size, $B$, thus consists of two terms: 1) a cross-entropy class prediction loss for the main task, and 2) a weighted binary cross-entropy loss for the oracle selection network. 
\begin{equation}
\begin{split}
\mathcal{L} = & \sum_{i=1}^{B} \overbrace{- \log\left (p(y_{i}=c|x_{i},\omega) \right)}^{\text{Class Prediction Loss}} - \overbrace{\beta e_{i}\log \left(g_{\theta}(o|x_{i}) \right) - (1-e_{i})\log \left(1-g_{\theta}(o|x_{i}) \right)}^{\text{Oracle Selection Loss}} \\
\end{split}
\label{eq:loss_function}
\end{equation}
where \textit{c} is the target class. To offset the aforementioned class imbalance, we introduce a dynamic hyperparameter, $\beta=\frac{\sum \delta_{e=0}}{\sum \delta_{e=1}}$, which changes according to the ratio of correctly classified to misclassified instances within a mini-batch, where $\delta$ is the Kronecker delta function. As training progresses, $\beta < 1 \rightarrow \beta > 1$. 

\begin{figure}[!h]
\centering
\begin{subfigure}[h]{0.32\textwidth}
        \centering
        \includegraphics[width=\textwidth]{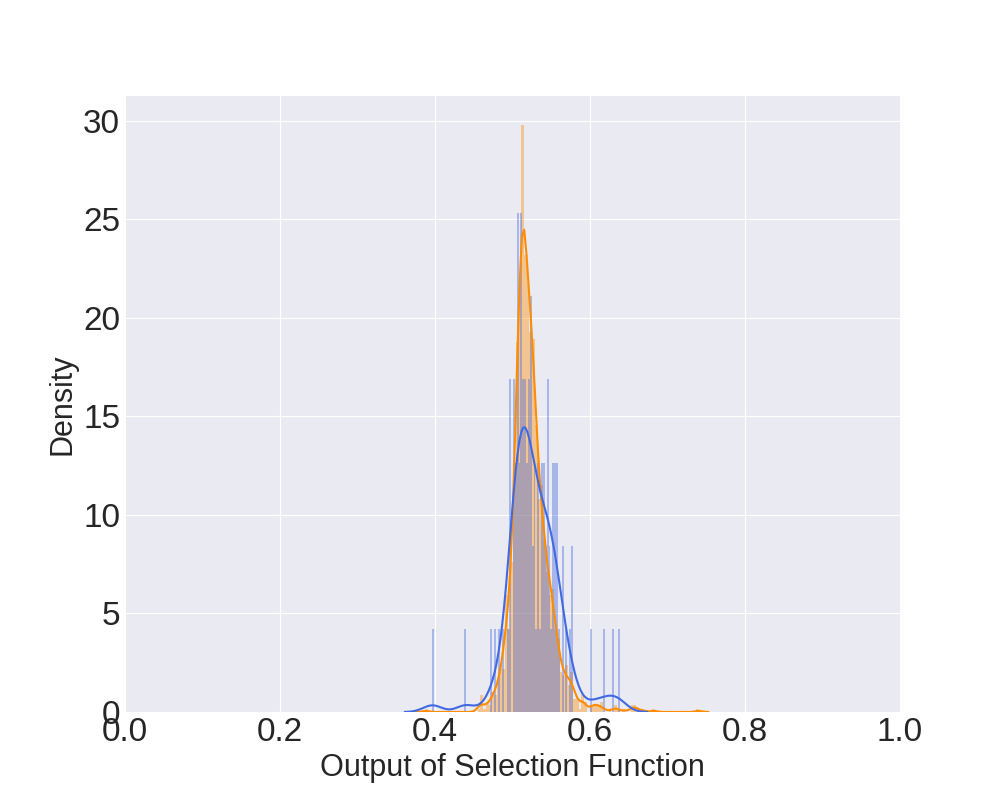}
        \caption{Early in Training}
\end{subfigure}
\begin{subfigure}[h]{0.32\textwidth}
        \centering
        \includegraphics[width=\textwidth]{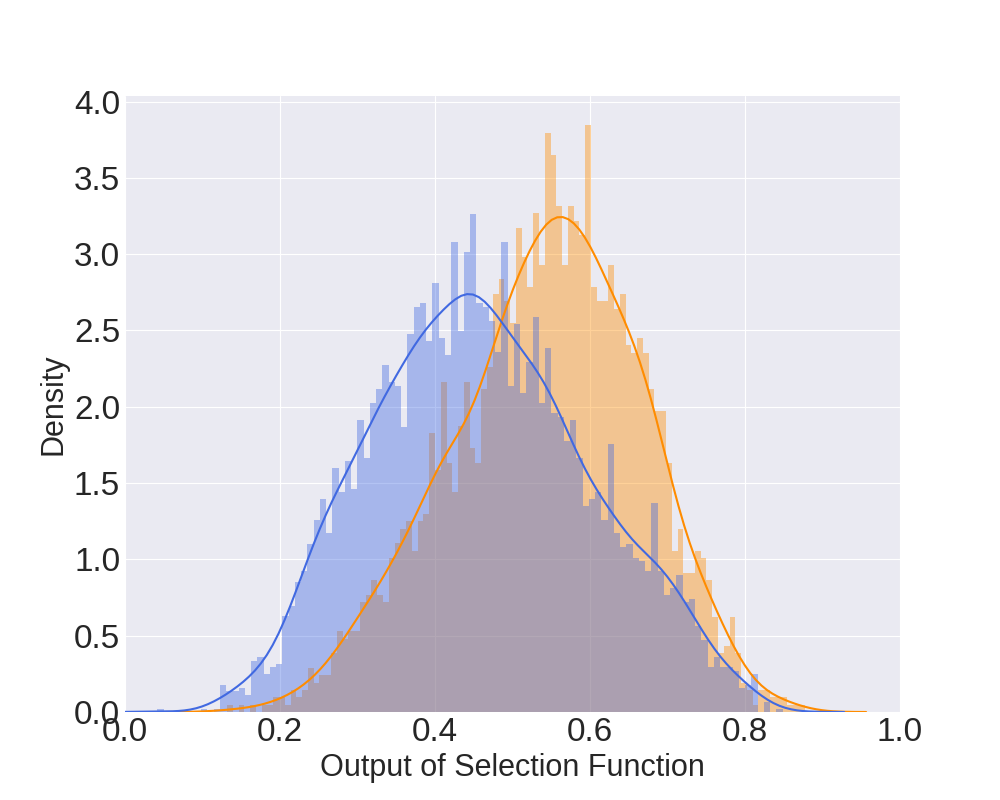}
        \caption{Late in Training}
        \label{fig:histogram_separation_late}
\end{subfigure}
\begin{subfigure}[h]{0.32\textwidth}
        \centering
        \includegraphics[width=\textwidth]{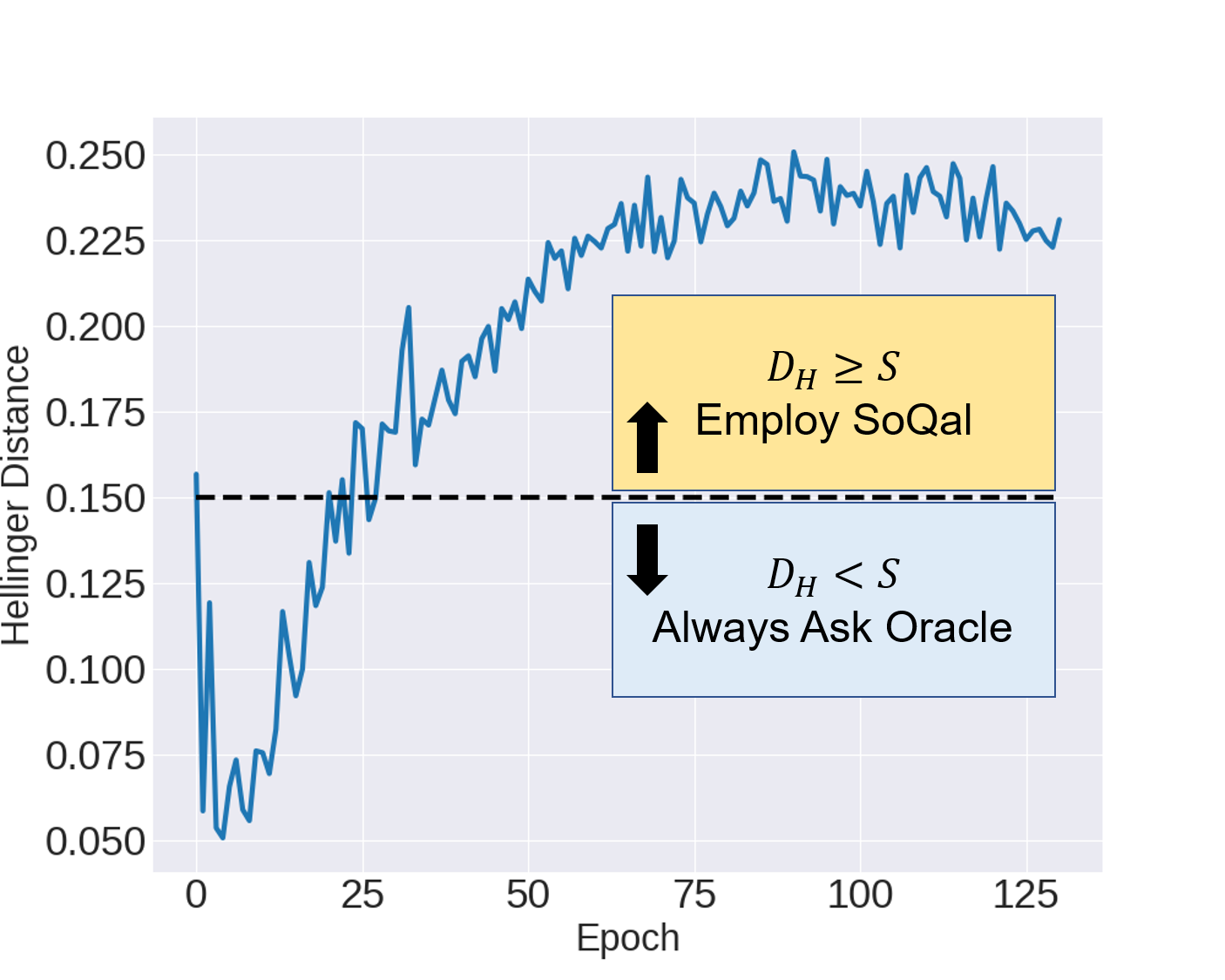}
        \caption{Hellinger Distance}
        \label{fig:hellinger_distance}
\end{subfigure}
        \caption{Density of the outputs of the oracle selection network $g_{\theta}$ 	conditioned on the zero-one classification error (a) early in training and (b) late in training. (c) Hellinger distance, $\mathcal{D}_{H}$, between distributions of outputs of selection function during training. Delegation of oracle questioning to the network occurs when $\mathcal{D}_{H} \geq S$. Notice the improved separability of the two distributions as a result of the training procedure.}
        \label{fig:histogram_separation}
\end{figure}

\textbf{Thresholding.} As we are dealing with unlabelled instances, we are interested in exploiting the output of $g_{\theta}$ as a proxy for whether an instance is correctly classified ($e=0$) or not ($e=1$). The separability of these two states determine the reliability of such a proxy. In Fig.~\ref{fig:histogram_separation_late}, we illustrate the distribution of the $o$ values that correspond to $e=0$ and $e=1$ on the \textit{labelled} training data.

At the end of each training epoch, the $o$ values in Fig.~\ref{fig:histogram_separation_late} were fit to two unimodal Gaussian distributions. This generates $\mathcal{N}_{0}(\mu_{0},\sigma_{0}^{2})$ and $\mathcal{N}_{1}(\mu_{1},\sigma_{1}^{2})$ for $e=0$ and $e=1$, respectively. We quantify the separability of these two distributions using the Hellinger distance, $\mathcal{D}_{H} \in [0,1]$. 
\begin{equation}
\mathcal{D}_{H} = \sqrt{1 - \sqrt{\frac{2\sigma_{0}\sigma_{1}}{\sigma_{0}^{2}\sigma_{1}^{2}}} e^{-\frac{1}{4}\frac{(\mu_{0} - \mu_{1})^{2}}{\sigma_{0}^{2}\sigma_{1}^{2}}}}
\end{equation}
If, at a particular acquisition epoch, $\mathcal{D}_{H}$ does not exceed some threshold \textit{S}, then $g_{\theta}$ cannot be relied upon and an oracle is always requested for a label. The value of \textit{S} can be altered depending on the degree of trust one has in the network and labeller. When $\mathcal{D}_{H} \geq S$, $\mathcal{N}_{0}$ and $\mathcal{N}_{1}$ are evaluated using the $o$ value for each acquired unlabelled instance. We outline the probability of asking an oracle, \textit{p}(A), in Eq.~\ref{eq:oracle_strategy}. Algorithms~\ref{algo:al} and \ref{algo:soqal} in Appendix~B illustrate the entire active learning procedure. 
\begin{equation}
p(\text{A}) = \begin{dcases}
1, & \mathcal{D}_{H} < S \\
1, & \mathcal{N}\left (o|\mu_{1}, \sigma_{1}^{2}, e=1\right) > \mathcal{N}\left (o|\mu_{0}, \sigma_{0}^{2}, e=0 \right) \text{and}\ \mathcal{D}_{H} \geq S\\
0, & \text{otherwise} \\
\end{dcases}
\label{eq:oracle_strategy}
\end{equation}

\subsection{Chernoff Bound on Error Rate of Selection Network}

Given that the selection network is tasked with making a binary decision, we can obtain a theoretical upper bound on its probability of making an error (via the overlap of density functions in Fig.~\ref{fig:histogram_separation}). An error in this context can be interpreted as stubbornness, where the network does not ask for help when it should have, and over-reliance, where the network asks for help when it should not have. The Chernoff upper bound on the error rate is as follows. The full derivation can be found in Appendix~C.
\begin{equation}
\begin{split}
P(\text{error})		
		& \leq P(e=0)^{\beta^{*}} P(e=1)^{1-\beta^{*}} e^{- \left[ \frac{\beta^{*}(1-\beta^{*})(\mu_{0} - \mu_{1})^{2}}{2(\beta^{*} \sigma_{0}^{2} + (1-\beta^{*})\sigma_{1}^{2})} + \frac{1}{2} log\frac{\beta^{*} \sigma_{0}^{2} + (1-\beta^{*})\sigma_{1}^{2}}{ \sigma_{0}^{2\beta^{*}}\sigma_{1}^{2(1-\beta^{*})}} \right]}
\end{split}
\label{eq:chernoff_bound}
\end{equation}
where $P(e=0)$ and $P(e=1)$ represent the prior probabilities of each class corresponding to the zero-one loss. $\beta^{*}$ is obtained by minimizing the exponent term.

\section{Experimental Design}

\subsection{Datasets}
\label{subsec:datasets}

Experiments were implemented in PyTorch \citep{Paszke2019} and were conducted on five publically-available datasets. These datasets consist of physiological time-series data such as the photoplethysmogram (PPG) and the electrocardiogram (ECG) alongside available cardiac arrhythmia labels. We use $\mathcal{D}_{1}$ = PhysioNet 2015 PPG, $\mathcal{D}_{2}$ = PhysioNet 2015 ECG \citep{Clifford2015} (5-way), $\mathcal{D}_{3}$ = PhysioNet 2017 ECG \citep{Clifford2017} (4-way), $\mathcal{D}_{4}$ = Cardiology ECG \citep{Hannun2019} (12-way), and $\mathcal{D}_{5}$ = PTB ECG \citep{Bousseljot1995} (2-way). 

\subsection{Baselines}

We experiment with baselines that exhibit varying degrees of oracle dependence. \textbf{No Oracle} was explored by \cite{Kiyasseh2020} where 0\% of label are oracle-based and are instead based on network predictions. \textbf{Epsilon Greedy} is a stochastic strategy from the reinforcement learning literature \cite{Watkins1989} where the degree of network exploration, performed with probability $\epsilon$, is decayed exponentially throughout training. In our case, we exponentially decay the reliance of the network on an oracle as a function of the number of acquisition epochs. \textbf{Entropy Response} assumes that high entropy predictions generated by a network are indicative of instances of which the network is unsure. Therefore, we introduce a threshold, $S_\mathrm{Entropy}$, such that if it is exceeded, an oracle is requested to label the instance chosen. The most dependent baseline is \textbf{100\% Oracle}, a traditionally-employed strategy in AL where 100\% of labels are oracle-based. 

We do not compare our methods to Softmax Response \citep{Geifman2017} and SelectiveNet \citep{Geifman2019}, despite their strong performance for selective classification, as they do not trivially extend to the setting in which labels are unavailable. 

\subsection{Hyperparameters}

\textbf{Active Learning.} For all experiments, we follow the hyperparameter choices in \cite{Kiyasseh2020}. Namely, we chose the number of MC samples \textit{T} = 20. Acquisitions of unlabelled instances were performed at pre-defined epochs of $\tau = 5n$, $n \in \mathbb{N}^{+}$. Moreover, the amount of instances acquired during each acquisition epoch is $b = 2\%$ of the remaining unlabelled instances. Lastly, we chose the temporal period $\Delta t$ = 1 for all experiments involving temporal variants of acquisition functions, as described later. 

\textbf{Selective Oracle Questioning.} In order for selective oracle questioning to be delegated to the network, we must have $\mathcal{D}_{H} \geq S$. Given that $\mathcal{D}_{H}$ was observed to have an increasing trend, as seen in Fig.~\ref{fig:hellinger_distance}, we chose $S=0.15$ to balance between reliability of the selection function and independence from an oracle. We also explore the sensitivity of SoQal to this choice of $S$.  

\section{Experiments}

\subsection{Selective Oracle Questioning with Noise-Free Oracle}
\begin{figure}[!h]
\centering
\begin{subfigure}[h]{0.9\textwidth}
        \centering
        \includegraphics[width=\textwidth]{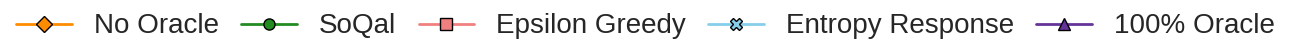}
\end{subfigure}
\begin{subfigure}[h]{0.38\textwidth}
        \centering
        \includegraphics[width=\textwidth]{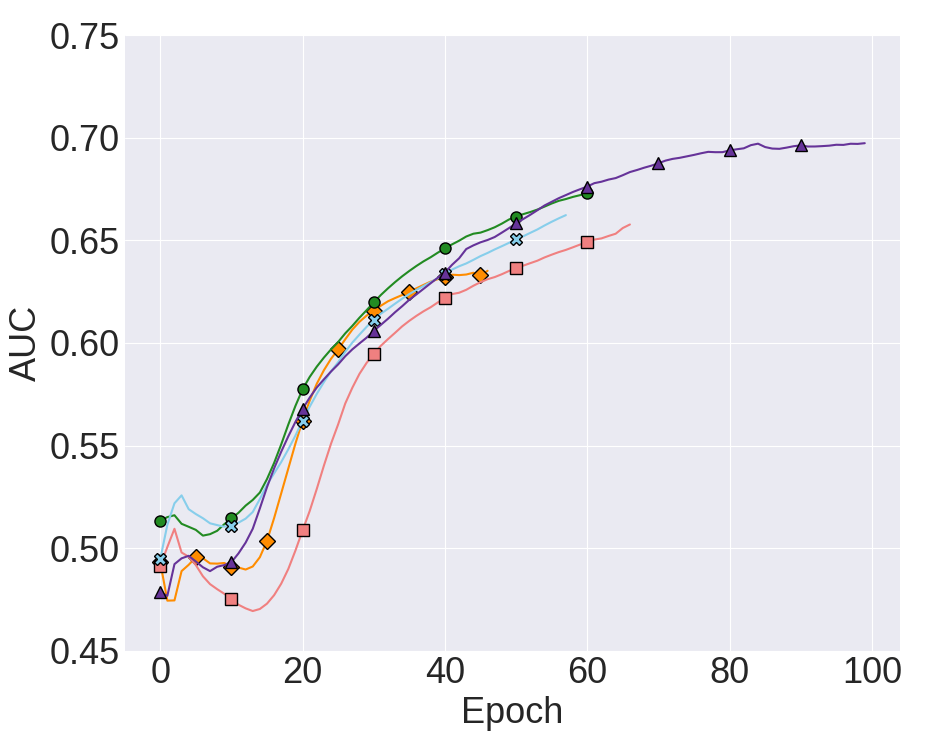}
        \caption{$\mathcal{D}_{2}$, BALD\textsubscript{MCP}}
        \label{fig:soq_1}
\end{subfigure}
\begin{subfigure}[h]{0.38\textwidth}
        \centering
        \includegraphics[width=\textwidth]{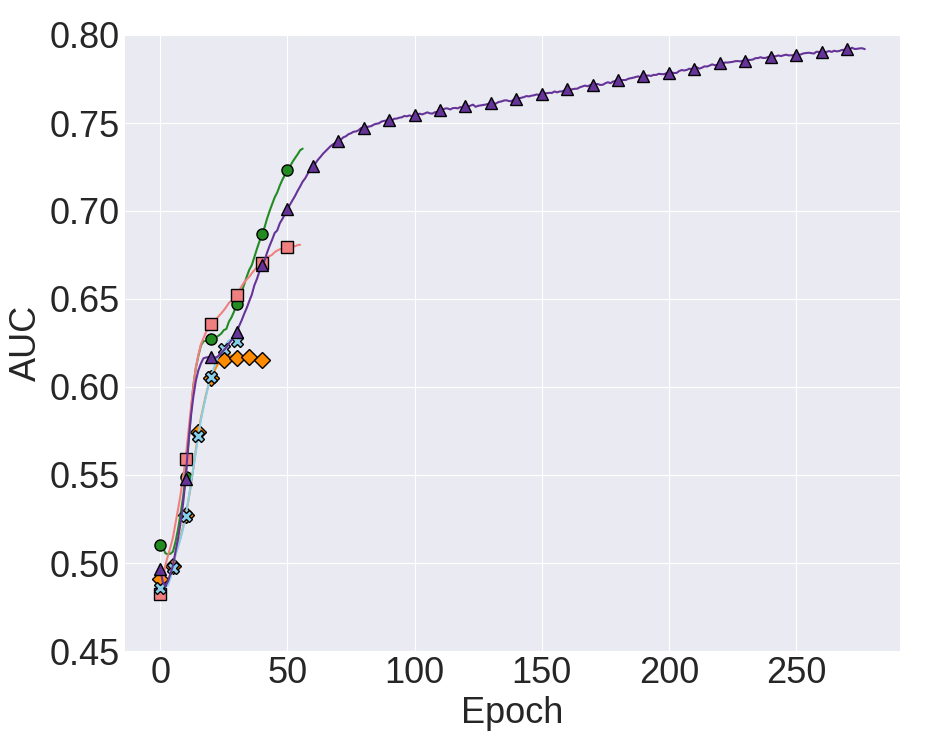}
        \caption{$\mathcal{D}_{3}$, BALD\textsubscript{MCD}}
        \label{fig:soq_2}
\end{subfigure}
        \caption{Mean validation AUC as a function of oracle selection strategies on (a) $\mathcal{D}_{2}$ using BALD\textsubscript{MCP} and (b) $\mathcal{D}_{3}$ using BALD\textsubscript{MCD}. Results are averaged across 5 seeds.}
        \label{fig:impact_of_soq_main}
\end{figure}

The ability of a learner to appropriately determine \textit{when} to request labels from an oracle can significantly alleviate the associated labelling burden. In this section, we evaluate this ability amongst the various oracle selection strategies. In Fig.~\ref{fig:impact_of_soq_main}, we illustrate the validation AUC of SoQal during training compared to that of the proposed baselines. We show that the 100\% oracle strategy outperforms the remaining methods. This can be seen by the $\mathrm{AUC}\approx0.70$ and $\approx0.80$ in Figs.~\ref{fig:soq_1} and \ref{fig:soq_2}, respectively. We expect this behaviour as labels from a noise-free oracle are likely to be accurate. Conversely, the no oracle strategy struggles, as seen by an $\mathrm{AUC}\approx0.64$ and $\approx0.62$ in Figs.~\ref{fig:soq_1} and \ref{fig:soq_2}, respectively. This can be explained by the idea that complete independence from an oracle, whereby labels are network-generated, is likely to lead to noisy labels and thus hinder performance. Based on these findings, it is clear that a dynamic oracle questioning strategy can offer a balance.

We illustrate, in Table~\ref{table:effect_of_oracle_strategies}, the test AUC of the oracle questioning strategies on all datasets. Across $\mathcal{D}_{1}$ - $\mathcal{D}_{3}$, we show that SoQal consistently outperforms its counterparts. For instance, while using BALD\textsubscript{MCD} on $\mathcal{D}_{2}$, SoQal achieves an $\mathrm{AUC}=0.707$ compared to $\mathrm{AUC}=0.609$ and $0.584$ for Epsilon Greedy and Entropy Response, respectively. These findings suggest that SoQal is better equipped to know \textit{when} and for which \textit{instance} a label is requested from an oracle. However, we observe that SoQal performs on par and relatively worse than the remaining methods on $\mathcal{D}_{4}$ and $\mathcal{D}_{5}$, respectively. We hypothesize that the former result is due to the cold-start problem \citep{Konyushkova2017} whereby AL algorithms fail to learn due to few available labelled training data. We support this claim with experiments in Appendix~I. As for the worse performance on $\mathcal{D}_{5}$, we believe this is due to the high degree of independence endowed upon the learner given the choice of $S$. Increasing the value of $S$ will cede control to the oracle and thus improve performance, an effect we quantify in Section~\ref{subsec:effect_of_S}. 

\definecolor{light-gray}{gray}{0.95}
\newcommand{\CC}[1]{\cellcolor{light-gray#1}}

\begin{table}[h]
\centering
\caption{Mean test AUC of oracle questioning strategies in the presence of a noise-free oracle. Results are shown for datasets $\mathcal{D}_{1} - \mathcal{D}_{5}$ and all acquisition functions. Mean and standard deviation values are shown across five seeds. 'No AL' is the strategy that does not employ active learning.}
\label{table:effect_of_oracle_strategies}
\vskip 0.1in 
\resizebox{\linewidth}{!}{
\begin{tabular}{c c | c c c c c c}
\hhline{========}
\multirow{2}{*}{Dataset} & \multirow{2}{*}{Ac. Function $\alpha$} & \multicolumn{5}{c}{Oracle Questioning Method} & \\ 
 &  & No Oracle & Entropy Response & Epsilon Greedy & SoQal (ours) & 100\% Oracle & No AL \\
\hhline{========}
\multirow{4}{*}{$\mathcal{D}_{1}$} & $\mathrm{BALD_{MCD}}$ & 0.465 $\pm$ 0.017 & 0.496 $\pm$ 0.039 & 0.491 $\pm$ 0.028 & \textbf{0.621 $\pm$ 0.021} & 0.653 $\pm$ 0.013 & \multirow{4}{*}{0.577 $\pm$ 0.014}\\
& $\mathrm{BALD_{MCP}}$ & 0.464 $\pm$ 0.023 & 0.517 $\pm$ 0.043 & 0.501 $\pm$ 0.043 & \textbf{0.645 $\pm$ 0.015} & 0.676 $\pm$ 0.020 & \\
& $\mathrm{BALC_{KLD}}$ & 0.500 $\pm$ 0.023 & 0.548 $\pm$ 0.034 & 0.548 $\pm$ 0.042 & \textbf{0.598 $\pm$ 0.055} & 0.634 $\pm$ 0.030 & \\ 
& Temporal $\mathrm{BALC_{KLD}}$ & 0.496 $\pm$ 0.024 & 0.536 $\pm$ 0.040 & 0.521 $\pm$ 0.059 & \textbf{0.646 $\pm$ 0.067} & 0.659 $\pm$ 0.033 & \\
\hline
\multirow{4}{*}{$\mathcal{D}_{2}$} & $\mathrm{BALD_{MCD}}$ & 0.573 $\pm$ 0.063 & 0.584 $\pm$ 0.041 & 0.609 $\pm$ 0.071 & \textbf{0.707 $\pm$ 0.038} & 0.713 $\pm$ 0.053 & \multirow{4}{*}{0.679 $\pm$ 0.040}\\
& $\mathrm{BALD_{MCP}}$ & 0.589 $\pm$ 0.045 & 0.638 $\pm$ 0.043 & 0.637 $\pm$ 0.044 & \textbf{0.677 $\pm$ 0.042} & 0.735 $\pm$ 0.028 & \\
& $\mathrm{BALC_{KLD}}$  & 0.602 $\pm$ 0.044 & 0.582 $\pm$ 0.017 & 0.643 $\pm$ 0.033 & \textbf{0.677 $\pm$ 0.024} & 0.722 $\pm$ 0.018 &\\ 
& Temporal $\mathrm{BALC_{KLD}}$ & 0.575 $\pm$ 0.017 & 0.612 $\pm$ 0.050 & 0.605 $\pm$ 0.019 & \textbf{0.648 $\pm$ 0.057} & 0.735 $\pm$ 0.011 &\\
\hline
\multirow{4}{*}{$\mathcal{D}_{3}$} & $\mathrm{BALD_{MCD}}$ & 0.581 $\pm$ 0.014 & 0.588 $\pm$ 0.013 & 0.673 $\pm$ 0.015 & \textbf{0.721 $\pm$ 0.025} & 0.802 $\pm$ 0.008 & \multirow{4}{*}{0.716 $\pm$ 0.012}\\
& $\mathrm{BALD_{MCP}}$  & 0.623 $\pm$ 0.020 & 0.676 $\pm$ 0.058 & 0.665 $\pm$ 0.028 & \textbf{0.720 $\pm$ 0.044} & 0.798 $\pm$ 0.007 & \\
& $\mathrm{BALC_{KLD}}$  & 0.631 $\pm$ 0.010 & 0.629 $\pm$ 0.004 & 0.643 $\pm$ 0.041 & \textbf{0.731 $\pm$ 0.033} & 0.787 $\pm$ 0.008 & \\ 
& Temporal $\mathrm{BALC_{KLD}}$ & 0.600 $\pm$ 0.005 & 0.630 $\pm$ 0.014 & 0.654 $\pm$ 0.019 & \textbf{0.730 $\pm$ 0.024} & 0.794 $\pm$ 0.002 &\\
\hline
\multirow{4}{*}{$\mathcal{D}_{4}$} & $\mathrm{BALD_{MCD}}$ & 0.486 $\pm$ 0.011 & 0.489 $\pm$ 0.030 & 0.474 $\pm$ 0.037 & 0.468 $\pm$ 0.021 & 0.585 $\pm$ 0.011 & \multirow{4}{*}{0.486 $\pm$ 0.023} \\
& $\mathrm{BALD_{MCP}}$ & 0.493 $\pm$ 0.030 & 0.504 $\pm$ 0.026 & 0.492 $\pm$ 0.024 & 0.499 $\pm$ 0.029 & 0.605 $\pm$ 0.024 &\\
& $\mathrm{BALC_{KLD}}$ & 0.505 $\pm$ 0.032 & 0.504 $\pm$ 0.039 & 0.473 $\pm$ 0.010 & 0.495 $\pm$ 0.012 & 0.588 $\pm$ 0.033 &\\ 
& Temporal $\mathrm{BALC_{KLD}}$ & 0.511 $\pm$ 0.030 & 0.496 $\pm$ 0.023 & 0.496 $\pm$ 0.023 & 0.503 $\pm$ 0.010 & 0.532 $\pm$ 0.027 &\\
\hline
\multirow{4}{*}{$\mathcal{D}_{5}$} & $\mathrm{BALD_{MCD}}$ & 0.717 $\pm$ 0.006 & 0.715 $\pm$ 0.005 & \textbf{0.718 $\pm$ 0.006} & 0.661 $\pm$ 0.105 & 0.937 $\pm$ 0.004 & \multirow{4}{*}{0.710 $\pm$ 0.097} \\
& $\mathrm{BALD_{MCP}}$ & 0.719 $\pm$ 0.009 & 0.678 $\pm$ 0.074 & \textbf{0.774 $\pm$ 0.047} & 0.453 $\pm$ 0.136 & 0.705 $\pm$ 0.013 & \\
& $\mathrm{BALC_{KLD}}$ & 0.679 $\pm$ 0.056 & 0.664 $\pm$ 0.064 & \textbf{0.726 $\pm$ 0.019} & 0.638 $\pm$ 0.145 & 0.900 $\pm$ 0.036 & \\ 
& Temporal $\mathrm{BALC_{KLD}}$ & 0.720 $\pm$ 0.010 & 0.689 $\pm$ 0.061 & \textbf{0.741 $\pm$ 0.028} & 0.571 $\pm$ 0.161 & 0.708 $\pm$ 0.002 &\\
\hhline{========}
\end{tabular}}
\end{table}

\subsection{Selective Oracle Questioning with Noisy Oracle}
\begin{figure}[!h]
    \centering
    \begin{subfigure}[h]{0.9\textwidth}
            \centering
            \includegraphics[width=\textwidth]{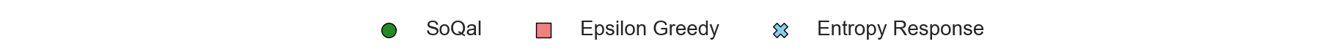}
    \end{subfigure}
    \begin{subfigure}[h]{0.75\textwidth}
            \centering
            \includegraphics[width=\textwidth]{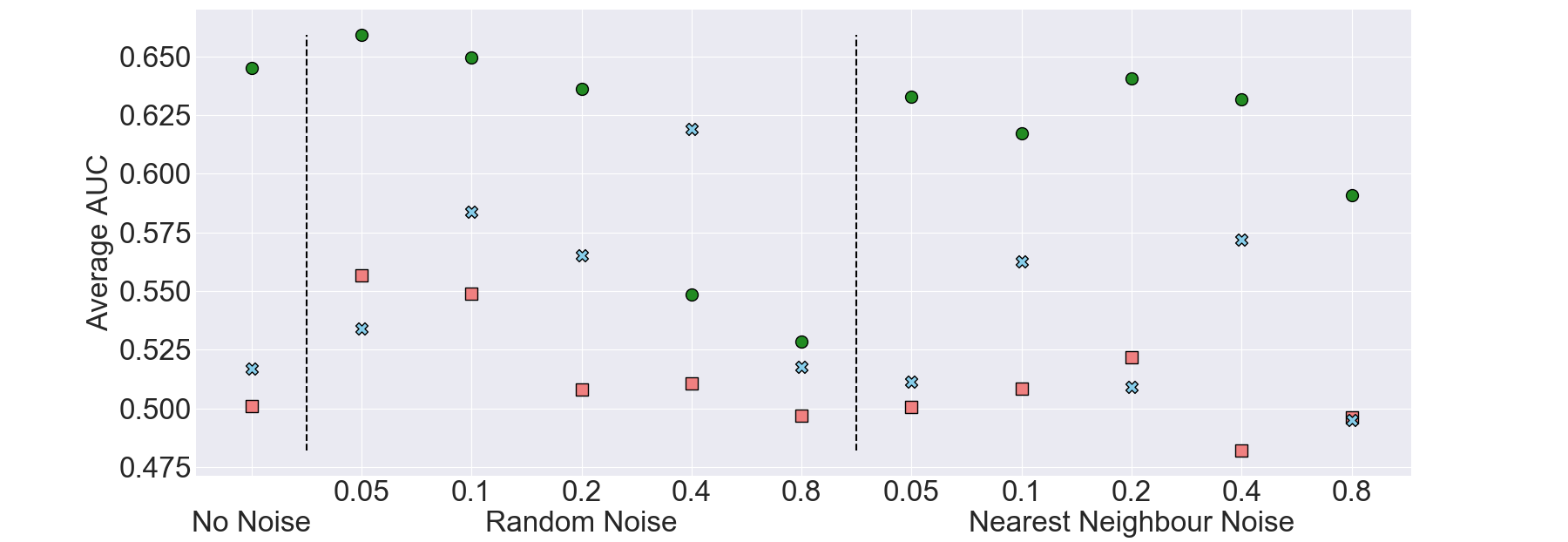}
            \caption{$\mathcal{D}_{1}$, $\mathrm{BALD_{MCP}}$}
            \label{fig:average_auc_physionetppg_baldmcp}
    \end{subfigure}	
	\caption{Average AUC of the oracle questioning strategies in the absence and presence of various magnitudes of label noise on $\mathcal{D}_{1}$ using $\mathrm{BALD_{MCP}}$. With up to 80\% random or nearest neighbour label noise, SoQal still outperforms its counterpart methods that are trained \textit{without} label noise.}
	\label{fig:average_auc_performance}
\end{figure}

In healthcare, physicians may be ill-trained, fatigued, or unable to diagnose a case due to its difficulty. We simulate these scenarios by introducing two types of label noise. We stochastically flip each label 1) to any other label randomly \textbf{(Random)}, or 2) to its nearest neighbour from a \textit{different} class in a compressed subspace \textbf{(Nearest Neighbour)}. Whereas the first form of noise is extreme, the latter form is more realistic as it may represent uncertainty in physician diagnoses. To simulate various magnitudes of noise, we chose the probability of introducing noise, $\gamma=[0.05,0.1,0.2,0.4,0.8]$. In Fig.~\ref{fig:average_auc_performance}, we illustrate the effect of label noise on the test AUC of oracle questioning strategies. The remaining results can be found in Appendix~H.

In Fig.~\ref{fig:average_auc_performance}, we show that SoQal outperforms the remaining strategies across all noise types and levels (except with 40\% random noise). For instance, with 5\% random noise, SoQal achieves an $\mathrm{AUC}\approx0.66$ compared to $\mathrm{AUC}\approx0.56$ and $\approx0.53$ for Epsilon Greedy and Entropy Response, respectively. Secondly, SoQal is better able to deal with label noise than its counterparts. Specifically, SoQal with 80\% random noise achieves $\mathrm{AUC}\approx0.53$ which is greater than $\mathrm{AUC}\approx0.50$ and $\approx0.52$ achieved by Epsilon Greedy and Entropy Response with no noise, respectively. This effect, which is even more pronounced when dealing with nearest neighbour noise, indicates the utility of SoQal in the presence of a noisy oracle. Finally, we observe that the introduction of label noise occasionally improves performance. This can be seen by the increase in SoQal's AUC from 0.64 (no noise) to 0.66 (5\% random noise). We hypothesize that this is due to inherent label noise in the public datasets. Therefore, by introducing further noise, we may be nudging these labels towards their ground truth values. 

\subsection{Degree of Dependence of SoQal on Oracle}

\begin{figure}[!h]
\centering
\begin{subfigure}[h]{0.75\textwidth}
        \centering
        \includegraphics[width=\textwidth]{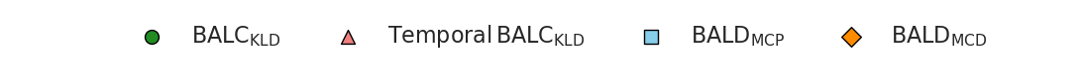}
\end{subfigure}
	\begin{subfigure}[h]{0.57\textwidth}
	        \centering
	        \includegraphics[width=\textwidth]{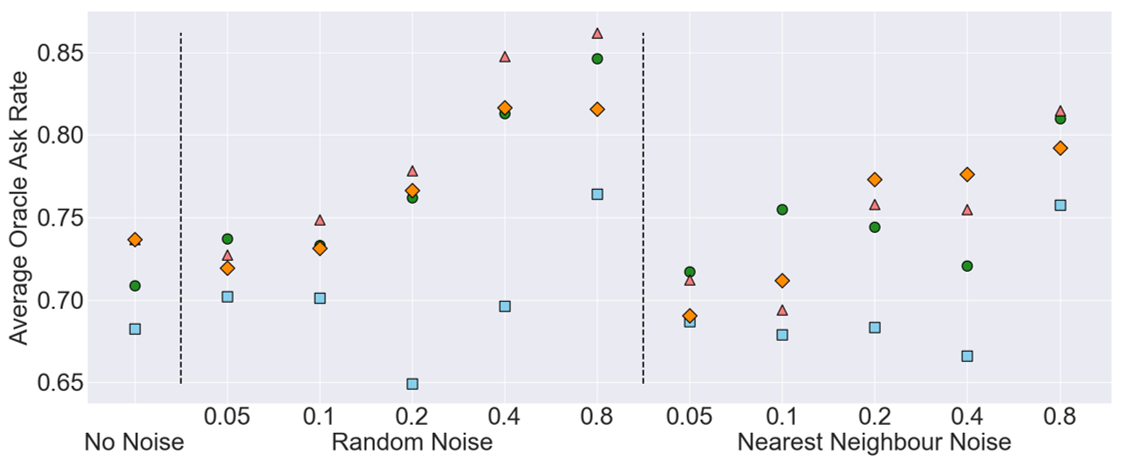}
	        \caption{Oracle ask-rate for different label noise scenarios}
	        \label{fig:average_oracle_ask_rate}
	\end{subfigure}
	\begin{subfigure}[h]{0.4\textwidth}
	        \centering
	        \includegraphics[width=\textwidth]{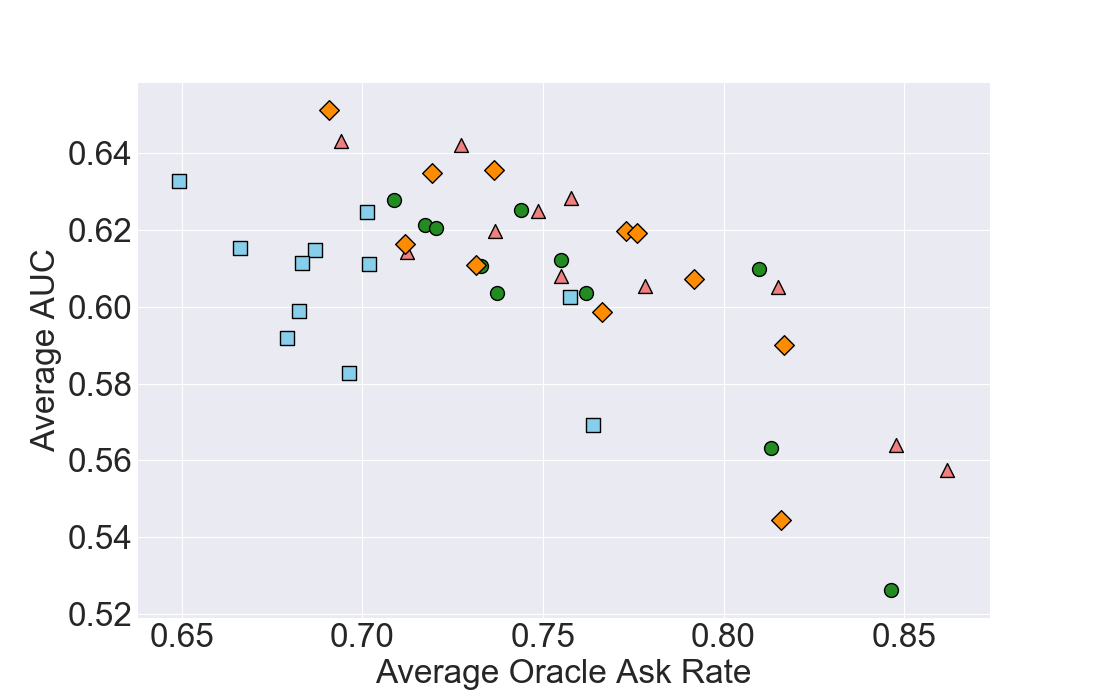}
	        \caption{Correlation between oracle ask-rate and generalization performance}
	        \label{fig:ask_rate_vs_performance}
	\end{subfigure}
	\caption{(a) SoQal's oracle ask-rate and (b) correlation between oracle ask-rate and average test AUC. Results are averaged across five seeds and all datasets, $\mathcal{D}_{1} - \mathcal{D}_{5}$, and are shown for each acquisition function and label noise scenario.}
	\label{fig:average_oracle_ask_rate_and_performance}
\end{figure}

It could be argued that the superiority of SoQal is simply due to high oracle dependence, as would be naively expected. In this section, we quantify SoQal's dependence on an oracle using the oracle ask-rate: the proportion of all instance acquisitions whose labels are requested from an oracle. In Fig.~\ref{fig:average_oracle_ask_rate}, we illustrate this oracle ask-rate for different label noise scenarios. 

In Fig.~\ref{fig:average_oracle_ask_rate}, we show that the oracle ask-rate varies based on the acquisition function used. For instance, at 20\% random noise, $\mathrm{BALD_{MCP}}$ requests labels 65\% of the time whereas the remaining acquisition functions do so approximately 77\% of the time. We hypothesize that this variability in the oracle ask-rate is due to the variability in the difficulty of the instances acquired by the acquisition functions. In other words, decreased dependence by $\mathrm{BALD_{MCP}}$ could be indicative of the acquisition of instances that are relatively farther away from the hyperplane. Thus, they are easier to classify and require less oracle guidance. 

In the presence of label noise, decreased dependence is indeed associated with improved generalization performance. This claim is supported by the negative correlation between the oracle ask-rate and the test AUC observed in Fig.~\ref{fig:ask_rate_vs_performance}. In other words, networks are requesting labels less and performing better. Such findings reaffirm the conclusion that SoQal knows \textit{when} to request a label from an oracle.  

\subsection{Controlling Oracle Dependence via Hellinger Threshold, $\pmb{S}$}
\label{subsec:effect_of_S}

When $\mathcal{D}_{H}<S$, all label requests are sent to the oracle. Therefore, the value of $S$ should control the oracle ask-rate and thus performance. We illustrate the performance of SoQal for a range of values of $S$ in Table~\ref{table:effect_of_S}. We confirm the expected positive relationship between $S$ and the oracle ask-rate where as $S=0.1 \rightarrow 0.4$, the oracle ask-rate increases from 86\% to 100\%. Moreover, for this particular dataset and acquisition function, $S=0.200$ is the optimal value as it achieves an $\mathrm{AUC}=0.768$ with an oracle ask-rate below 100\%. This finding reaffirms our previous hypothesis that the original labels in the dataset may be noisy. Therefore, not requesting these particular labels from the oracle is advantageous. 

\begin{table}[h]
\small
\centering
\caption{Mean test AUC of SoQal and oracle ask rate in response to various threshold values, $S$. Results are shown for $\mathcal{D}_{3}$ and BALD\textsubscript{MCD} across five seeds. Experiments are performed with a noise-free oracle.}
\label{table:effect_of_S}
\vskip 0.1in 
\begin{tabular}{c | c c c c c c c}
\hhline{========}
Threshold, $S$ & 0.100 & 0.125 & 0.150 & 0.175 & 0.200 & 0.300 & 0.400 \\
\hline
Average Oracle Ask Rate \% & 86 & 85 & 89 & 90 & 94 & 100 & 100 \\
AUC & 0.716 & 0.744 & 0.721 & 0.753 & 0.768 & 0.743 & 0.755 \\
\hhline{========}
\end{tabular}
\end{table} 

\section{Discussion and Future Work}

In this work, we proposed a dynamic oracle questioning strategy, SoQal, in the context of active learning and healthcare. We showed that while striking a balance between independence from and over-reliance on an oracle, SoQal outperforms strong baseline methods. Furthermore, in the presence of noisy oracles which represent ill-trained or fatigued physicians, SoQal decreases its dependence by 35\% and continues to outperform its counterparts. Indeed, we showed that this decreased dependence was appropriate and was correlated with improved generalization performance. We now mention several exciting avenues worth exploring. 

\textbf{Incorporating Prior Information.} As it stands, and in the absence of a priori information, the default mode for SoQal is deferral to an oracle. If, however, relevant a priori information is available (such as the extent of the noise inherent in the oracle's labels), then either the default mode or the Hellinger threshold, $S$, can be altered accordingly. The latter can also be changed during training if a dynamic label noise detector is present. 

\textbf{Incorporating Multiple Oracles.} In this work, we explored a dynamic oracle questioning strategy in the presence of a single oracle. Realistic clinical scenarios may include multiple experts of various levels of competency. Therefore, an interesting line of research could focus on a strategy that dynamically determines the \textit{ideal} expert for the instance at hand. 

\section{Broader Impact}

The exploration of less burdensome active learning algorithms in the context of healthcare can alleviate the exigent burden placed on medical practitioners. This is particularly acute in environments where burnout of physicians and nurses is increasingly observed due to a lack of electronic health record usability \citep{Melnick2020} and increased expectations \citep{Ferket2020}. The decreased dependence of SoQal on an expert, at appropriate times, could prevent the disruption of clinical workflows and thus put patient care at centre stage. On the other hand, inappropriate independence of an algorithm from an expert can lead to incorrect learning signals that are reinforced throughout the algorithm's lifetime. Consequently, a high degree of misdiagnoses can occur, thus negatively impacting clinical decision making and patient outcomes. SoQal attempts to balance this independence against diagnostic accuracy. 

As for low-resource clinical settings where physicians are either ill-trained or unavailable and labelled data is scarce, SoQal offers a scalable approach for operating in such environments. It labels data iteratively without over-dependence on potentially unreliable `experts'. This, in turn, generates large, labelled datasets that can be successfully leveraged by data-hungry deep learning algorithms. 

\bibliography{SoQal}
\bibliographystyle{unsrtnat}

\end{document}